\documentclass{article}

\usepackage[square,sort,comma,numbers]{natbib}
\usepackage[final]{nips_2019}

\usepackage[utf8]{inputenc} 
\usepackage[T1]{fontenc}    
\usepackage{hyperref}       
\usepackage{url}            
\usepackage{booktabs}       
\usepackage{amsfonts}       
\usepackage{nicefrac}       
\usepackage{microtype}      
\usepackage{changepage}
\usepackage{soul}

\usepackage{subcaption}
\usepackage{tikz,pgf,pgfplots}
\usepackage{pgfplotstable}
\usetikzlibrary{intersections}

\usepgfplotslibrary{fillbetween}
\usetikzlibrary{decorations.text}
\usetikzlibrary{positioning}

\tikzset{
    state/.style={
           rectangle,
           rounded corners,
           draw=black, very thick,
           minimum height=2em,
           inner sep=2pt,
           text centered,
           },
}

\usepackage{amsmath,amssymb}

\title{Using Generative Adversarial Networks to Synthesize Artificial Financial Datasets}

\author{
  Dmitry Efimov \\
  Credit and Fraud Risk\\
  American Express\\
  200 Vesey St\\
  New York, NY 10281 \\
  \texttt{dmitry.efimov@aexp.com} \\
  \And
  Di Xu \\
  Credit and Fraud Risk\\
  American Express\\
  200 Vesey St\\
  New York, NY 10281 \\
  \texttt{di.xu@aexp.com} \\  
  \AND
  Luyang Kong \\
  Credit and Fraud Risk\\
  American Express\\
  200 Vesey St\\
  New York, NY 10281 \\
  \texttt{luyang.kong1@aexp.com} \\  
  \And
  Alexey Nefedov \\
  Credit and Fraud Risk\\
  American Express\\
  200 Vesey St\\
  New York, NY 10281 \\
  \texttt{alexey.nefedov@aexp.com} \\
  \And
  Archana Anandakrishnan \\
  Credit and Fraud Risk\\
  American Express\\
  200 Vesey St\\
  New York, NY 10281 \\
  \texttt{archana.anandakrishnan@aexp.com} \\
}
\setcitestyle{square}
\pgfplotsset{compat=1.14}
\begin{document}

\maketitle

\begin{abstract}

Generative Adversarial Networks (GANs) became very popular for generation of realistically looking images. In this paper, we propose to use GANs to synthesize artificial financial data for research and benchmarking purposes. We test this approach on three American Express datasets, and show that properly trained GANs can replicate these datasets with high fidelity. For our experiments, we define a novel type of GAN, and suggest methods for data preprocessing that allow good training and testing performance of GANs. We also discuss methods for evaluating the quality of generated data, and their comparison with the original real data.

\end{abstract}

\section{Introduction}

Machine learning (ML) algorithms are ubiquitous in many practical domains, including the banking and financial industry, where some of their core applications are assessing creditworthiness of customers, offering customers optimal financial products, and identifying fraud. Measuring performance of these algorithms is critical to understanding their strengths and weaknesses. Comprehensive comparisons of algorithms require high quality datasets that can be used as standard benchmarks.

An integral part of the modern machine learning field is a corpus of publicly shared, high quality datasets from numerous domains, which are used to benchmark and validate ML algorithms developed by various researchers, academic groups and companies (see, for example, MNIST \cite{mnist1998}, ImageNet \cite{imagenet}, Kaggle \cite{kaggledatasets}, University of Irvine Machine Learning Repository \cite{dua2017}, etc. \cite{wiki}). These datasets also play an important role in advancing state of the field by greatly facilitating testing and development of novel ideas and methods.

In the long list of publicly available datasets, one can notice the shortage of datasets originating from the banking and financial industry, especially datasets associated with credit and fraud risk management operations. One of the key reasons for this is that making such data publicly available would be a violation of customers' privacy and trust.

For several available financial datasets (e.g. \cite{findata1}, \cite{findata2}), this issue is resolved by some pre-treatment of data, for example by reducing data to principal components, or by taking only a small sample of data \cite{dal2015}, \cite{yeh2009}. However, these transformed or subsampled datasets may fail to capture some unique properties of financial data. For example, transactional data is usually large, highly structured, contains a wide range of both numerical (continuous and discrete) and categorical variables, variables with very skewed distributions, missing values, etc. Practical algorithms should be able to deal with such variables in a robust and efficient manner, and be scalable to the large size of this data. To develop state of the art ML methods, including methods for anomaly detection and model interpretation, ML researchers and practitioners need to have access to data that is as close to the real one as possible.

In this paper, we suggest using Generative Adversarial Networks (GANs) as a means to create synthetic, ``artificial'' financial data. We describe experiments with three American Express datasets used for risk modeling. We show that properly trained GANs can replicate these datasets with high fidelity. Specifically, we show that synthesized data follows the same distribution as the original data, and that ML models trained on synthesized data have the same performance as those trained on the original data. In our experiments, we use a novel type of GAN architecture combining conditional GAN and DRAGAN, which gives us better training convergence and testing performance.

Since GAN-generated data does not originate from real customers, it could be made public as a benchmark dataset for financial applications. Customers' personal data and privacy would be protected using this approach.

This paper is structured as follows. Section \ref{Sec:GAN} gives a brief introduction to GANs and introduces a novel type of GAN that we used in our experiments. In Section \ref{Sec:processing}, we discuss data preprocessing methods that can help to improve training and testing performance of GAN. In Section \ref{Sec:evaluation}, we discuss methods for evaluation of the generated datasets and illustrate that any improvements made using the generated datasets scale and generalize to the real dataset. We conclude by discussing future steps in our research and other potential applications of GANs in the financial industry.

\section{Data Generation with Generative Adversarial Networks}
\label{Sec:GAN}

Generative Adversarial Network (GAN) is a type of neural network comprised of two connected networks, called generator and discriminator, which compete with each other during the training phase (\cite{goodfellow2014}). The objective of the discriminator is to distinguish samples coming from a given training set (“real data”) from samples created by generator (“synthesized data”). The error of discriminator is fed to the generator, which learns to produce samples that are increasingly difficult for discriminator to distinguish from real ones.

When training of GAN is complete, its generator can be used to synthesize data reproducing original, real data that was used for training. Since its introduction in 2014, GANs have mostly been used to generate realistically looking images in image classification and computer vision applications \cite{perez2017}, \cite{shin2018}, \cite{zhu2017}. However, there has been an increasing number of GAN applications to non-image data, where the goal of GANs can be, for example, enhancing real training data with synthetic samples \cite{zheng2018}, \cite{fiore2017}, \cite{kumar2018}.

In our experiments, the primary goal of using GANs was to generate synthetic data that replicate distribution of original real data, and allow us to build ML models that perform on par with ML models built using original real data. We used three American Express datasets from three different use cases in risk modeling. Basic details of these datasets are given in Table \ref{tab:datasets}. The datasets were chosen to represent some variety of data features, feature distributions, and type of classification problems. We omit further details about our use cases; however, they are not relevant in the scope of this paper. 

\begin{table*}[h]
  \begin{tabular}{lccr}
    \hline
    Dataset & Number of Features (Numerical, Categorical) & Size of Data Sample & Target Variable \\
    \hline
 Dataset A & 22 (21,1) & \;\,\,120,990 & Continuous \\
 Dataset B & 119 (113, 6) & 2,197,762 & Binary \\
 Dataset C & 471 (407, 64)  & 2,028,106 & Binary \\
 \hline
  \end{tabular}
  \caption{\label{tab:datasets} Details of three datasets used in our experiments with GANs.}
\end{table*}

In the following paragraphs, we are going to give basic definitions related to GANs, and introduce a new type of GAN that we used in our experiments.

In a classic (vanilla) GAN, generator network $G$ takes a random noise vector $\mathbf{z}$ as input, and produces fake sample $G_{\varphi}(\mathbf{z})$, where $\varphi$ is a set of generator's parameters (Fig. \ref{fig:1} (left)). Discriminator network $D$ takes in a sample $\mathbf{x}$, which can be either real or fake, and computes probability $D_{\theta}(\mathbf{x})$, where $\theta$ is a set of discriminator's parameters. The goal of discriminator is to distinguish between real and fake (generated) samples:
\begin{equation}
D_{\theta}(\mathbf{x}) = \left\{\begin{array}{ll}
1, & \mbox{if } \mathbf{x} \mbox{ is a real sample}, \\
0, & \mbox{if } \mathbf{x} \mbox{ is a generated sample}
\end{array}\right.
\end{equation}

GAN training is performed by sequential minimization of discriminator's loss function $J^{(D)} (\varphi, \theta)$ with respect to parameters $\theta$, where $J^{(D)} (\varphi, \theta)$ is defined as
\begin{equation}
J^{(D)} (\varphi, \theta) = -\mathbb{E}_{\mathbf{x}} \log D_{\theta} (\mathbf{x}) - \mathbb{E}_{\mathbf{z}} \log (1 - D_{\theta} (G_{\varphi}(\mathbf{z}))),
\end{equation}
and minimization of generator's loss function $J^{(G)} (\varphi, \theta)$ with respect to parameters $\varphi$, where $J^{(G)} (\varphi, \theta)$ is defined as
\begin{equation}
J^{(G)} (\varphi, \theta) = - \mathbb{E}_{\mathbf{z}} \log D_{\theta} (G_{\varphi}(\mathbf{z})).
\end{equation}

Training of vanilla GAN may be unstable due to gradient exploding or gradient vanishing effects \cite{gulrajani2017}. These effects can be even stronger in applications with non-image, structured data.

To overcome this problem, Deep Regret Analytic Generative Adversarial Networks (DRAGANs) were introduced by Kodali et al. in \cite{kodali2017}. In DRAGAN, discriminator's loss is modified by adding a regularization term that prevents sharp gradients and improves convergence:
\begin{equation}
J^{(D)}_{*} (\varphi, \theta) = J^{(D)} (\varphi, \theta) + \lambda \cdot \mathbb{E}_{\mathbf{x}, \delta \sim N_d (0, cI)} \left[|| \nabla_{\mathbf{x}} D_{\theta} (\mathbf{x} + \delta) || - k\right]^{2},
\end{equation}
where $\lambda$, $c$ and $k$ are hyperparameters. Fig. \ref{fig:2} shows the difference in convergence of regular GAN and DRAGAN on our data; the convergence of DRAGAN is much more stable.

{Conditional GANs (CGANs) were introduced by Mirza et al. in \cite{mirza2014} to better handle categorical features in training data. In CGAN, generator's input contains two parts: a random noise vector $\mathbf{z}$ and a dummy features vector $\mathbf{y}$ generated from categorical features (Fig. \ref{fig:1} (right)). Generator's output consists of vector $\mathbf{x}$ of numerical features only. Before feeding to the discriminator, vectors $\mathbf{y}$ and $\mathbf{x}$ are concatenated and equations (2) and (3) are modified as
\begin{equation}
J^{(G)} (\varphi, \theta) = - \mathbb{E}_{\mathbf{z}} \log D_{\theta} (G_{\varphi}(\mathbf{z}, \mathbf{y}), \mathbf{y})
\end{equation}
and
\begin{equation}
J^{(D)} (\varphi, \theta) = -\mathbb{E}_{\mathbf{x}, \mathbf{y}} \log D_{\theta} (\mathbf{x}, \mathbf{y}) - \mathbb{E}_{\mathbf{z}} \log (1 - D_{\theta} (G_{\varphi}(\mathbf{z}, \mathbf{y}), \mathbf{y}))
\end{equation}
accordingly. 

Since our data contained categorical features, we decided to use a combination of DRAGAN with CGAN architecture, which we called conditional DRAGAN (CDRAGAN). In CDRAGAN, we add regularization term from (4) to the discriminator loss  (6) in order to avoid exploding gradients. Just like in non-conditional case, convergence of CDRAGAN has better stability than convergence of CGAN (Fig. \ref{fig:3}).

\begin{figure}
\resizebox{0.5\textwidth}{!}{
\begin{tikzpicture}

	\node (z) {$\mathbf{z}$};
	\node[circle, draw, thick, right=2em of z] (G) {$G_{\varphi}(\mathbf{z})$};
	\draw[-stealth, thick] (z) -- (G);
	\node[right=2em of G] (x) {$\mathbf{x}_{fake}$};
	\draw[-stealth, thick] (G) -- (x);
	\node[above=of x] (xt) {$\mathbf{x}_{real}$};
	\node[right=2em of x, circle, fill, inner sep=0.15em] (pt1) {};
	
	\node[right=2em of xt, circle, fill, inner sep=0.15em] (pt2) {};
	\node[thick, fill=white, inner sep=0.15em] at ([xshift=-0.1em, yshift=1.5em]pt1.north) (pt3) {$\mathbf{x}$};
	\node[circle, draw, thick, right=4em of x, yshift=2.5em] (D) {$D_{\theta}(\mathbf{x})$};
	
	\node[left=10em of xt] (it) {};
	\node[right=2em of D] (out) {real?};
	\draw[dashed, thick] (pt1) edge[bend left] (pt2);
	\draw[-stealth, thick] (x) -- (pt1);
	\draw[-stealth, thick] (xt) -- (pt2);
	\draw[-stealth, thick] (pt3) -- (D);
	\draw[-stealth, thick] (D) -- (out);

\end{tikzpicture}
}
\resizebox{0.5\textwidth}{!}{
\begin{tikzpicture}

	\node (y) {$\mathbf{y}$};
	\node[below=1em of y] (z) {$\mathbf{z}$};
	\node[circle, draw, thick, right=2em of y, yshift=-1em] (G) {$G_{\varphi}(\mathbf{z}, \mathbf{y})$};
	\draw[-stealth, thick] (z) -- (G);
	\draw[-stealth, thick] (y) -- (G);
	
	\node[right=2em of G] (x) {$\mathbf{x}_{fake}$};
	\draw[-stealth, thick] (G) -- (x);
	\node[above=of x] (xt) {$\mathbf{y}, \mathbf{x}_{real}$};
	\node[right=2em of x, circle, fill, inner sep=0.15em] (pt1) {};
	
	\node[right=2em of xt, circle, fill, inner sep=0.15em] (pt2) {};
	\node[thick, fill=white, inner sep=0.15em] at ([xshift=0.4em, yshift=1.4em]pt1.north) (pt3) {$\mathbf{y}, \mathbf{x}$};
	\node[circle, draw, thick, right=5em of x, yshift=2.5em] (D) {$D_{\theta}(\mathbf{x}, \mathbf{y})$};
	
	\node[left=10em of xt] (it) {};
	\node[right=1.5em of D] (out) {real?};
	\draw[dashed, thick] (pt1) edge[bend left] (pt2);
	\draw[-stealth, thick] (x) -- (pt1);
	\draw[-stealth, thick] (xt) -- (pt2);
	\draw[-stealth, thick] (pt3) -- (D);
	\draw[-stealth, thick] (D) -- (out);
	\draw[-stealth, thick] (y) edge[bend left] (x);

\end{tikzpicture}
}
\caption{Comparing architectures of GAN (left) and conditional GAN (right)} \label{fig:1}
\end{figure}
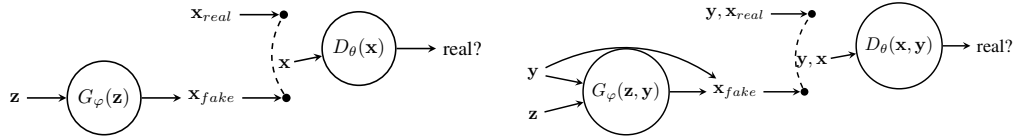

\begin{figure}[h]
\begin{subfigure}{0.5\textwidth}
\includegraphics[width=\linewidth]{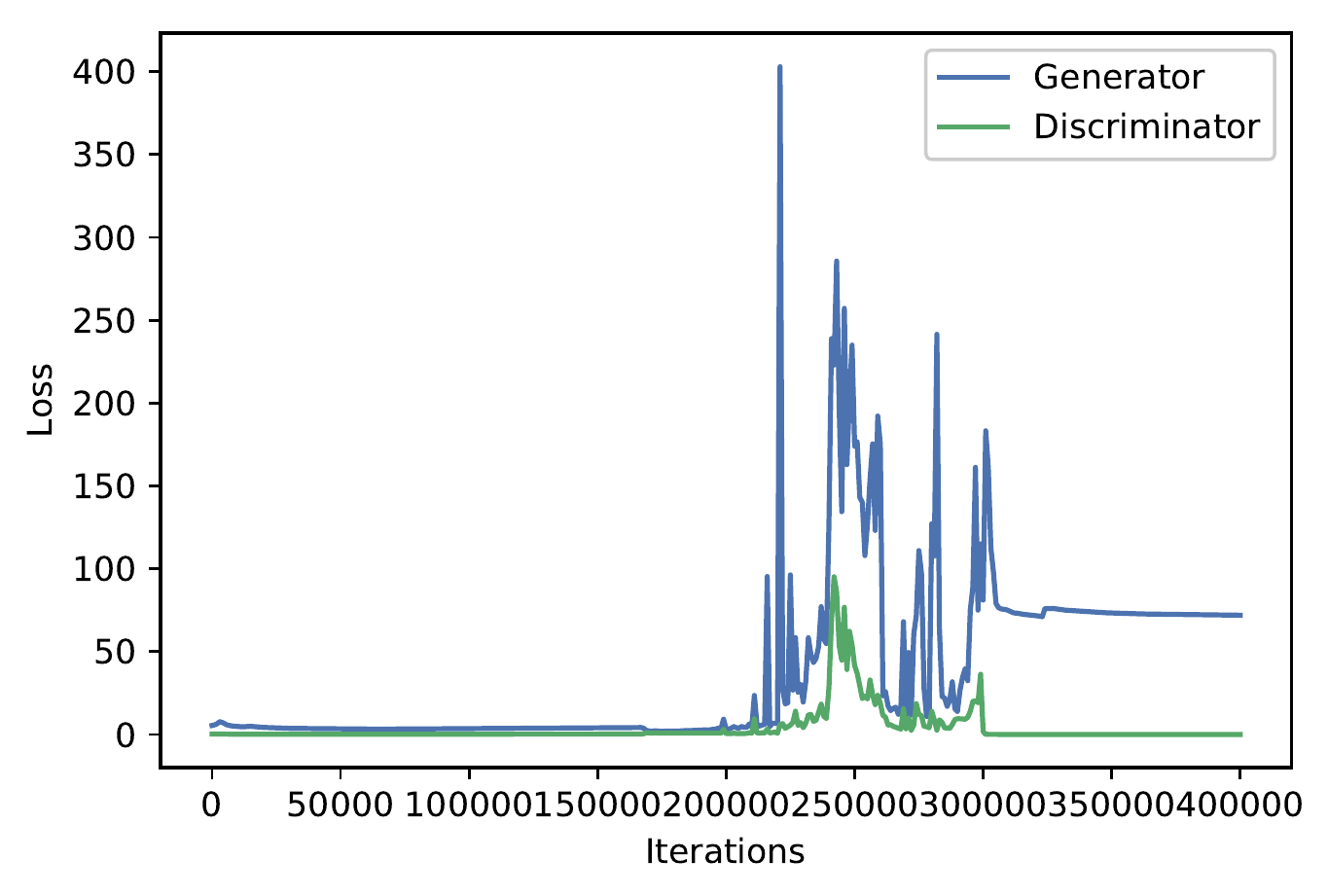}
\caption{Vanilla GAN} \label{fig:2a}
\end{subfigure}
\hspace*{\fill}
\begin{subfigure}{0.5\textwidth}
\includegraphics[width=\linewidth]{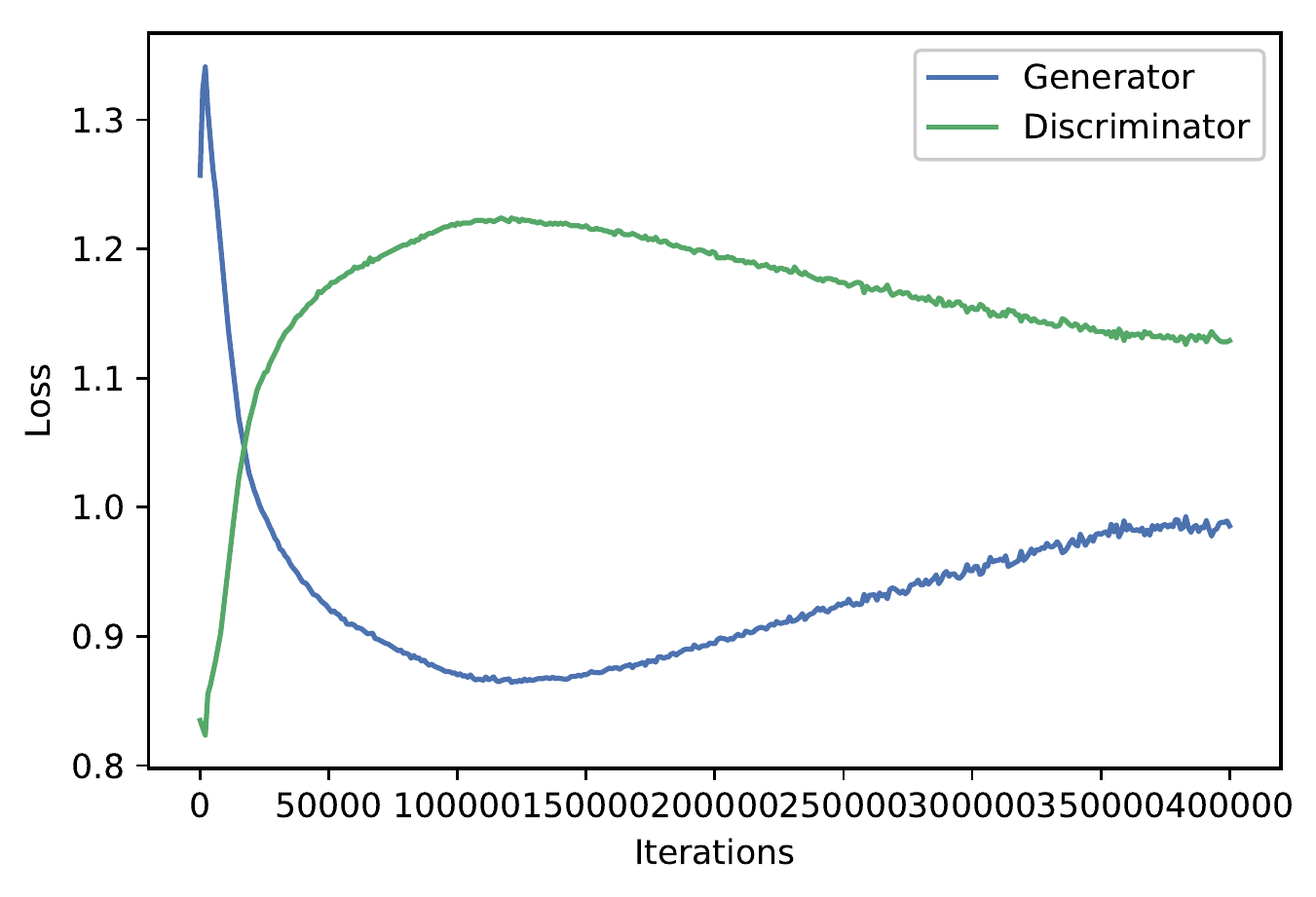}
\caption{DRAGAN} \label{fig:2b}
\end{subfigure}
\caption{Comparing convergence of vanilla GAN and DRAGAN} \label{fig:2}
\end{figure}

\begin{figure}[h]
\begin{subfigure}{0.5\textwidth}
\includegraphics[width=\linewidth]{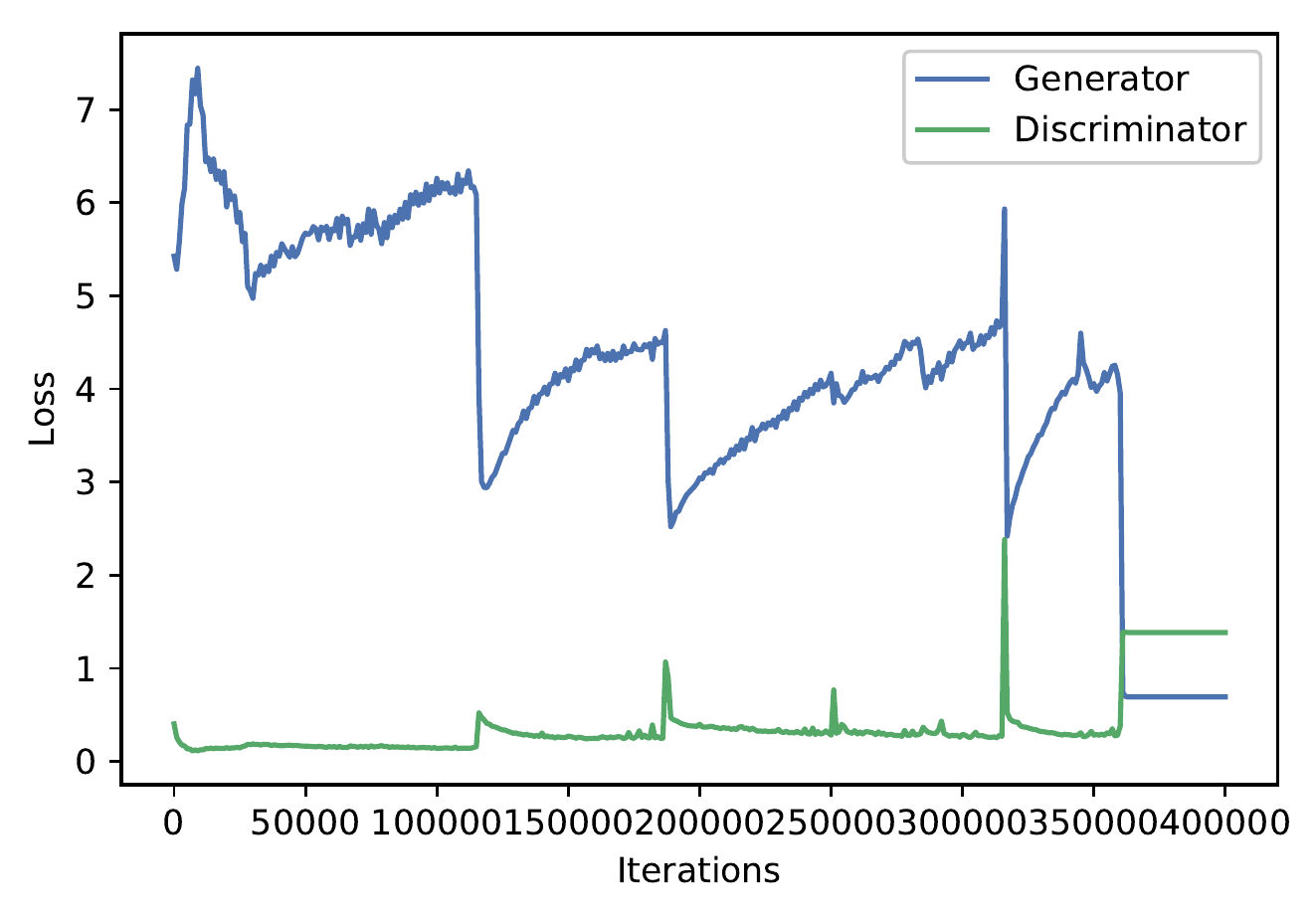}
\caption{Conditional GAN} \label{fig:3a}
\end{subfigure}
\hspace*{\fill}
\begin{subfigure}{0.5\textwidth}
\includegraphics[width=\linewidth]{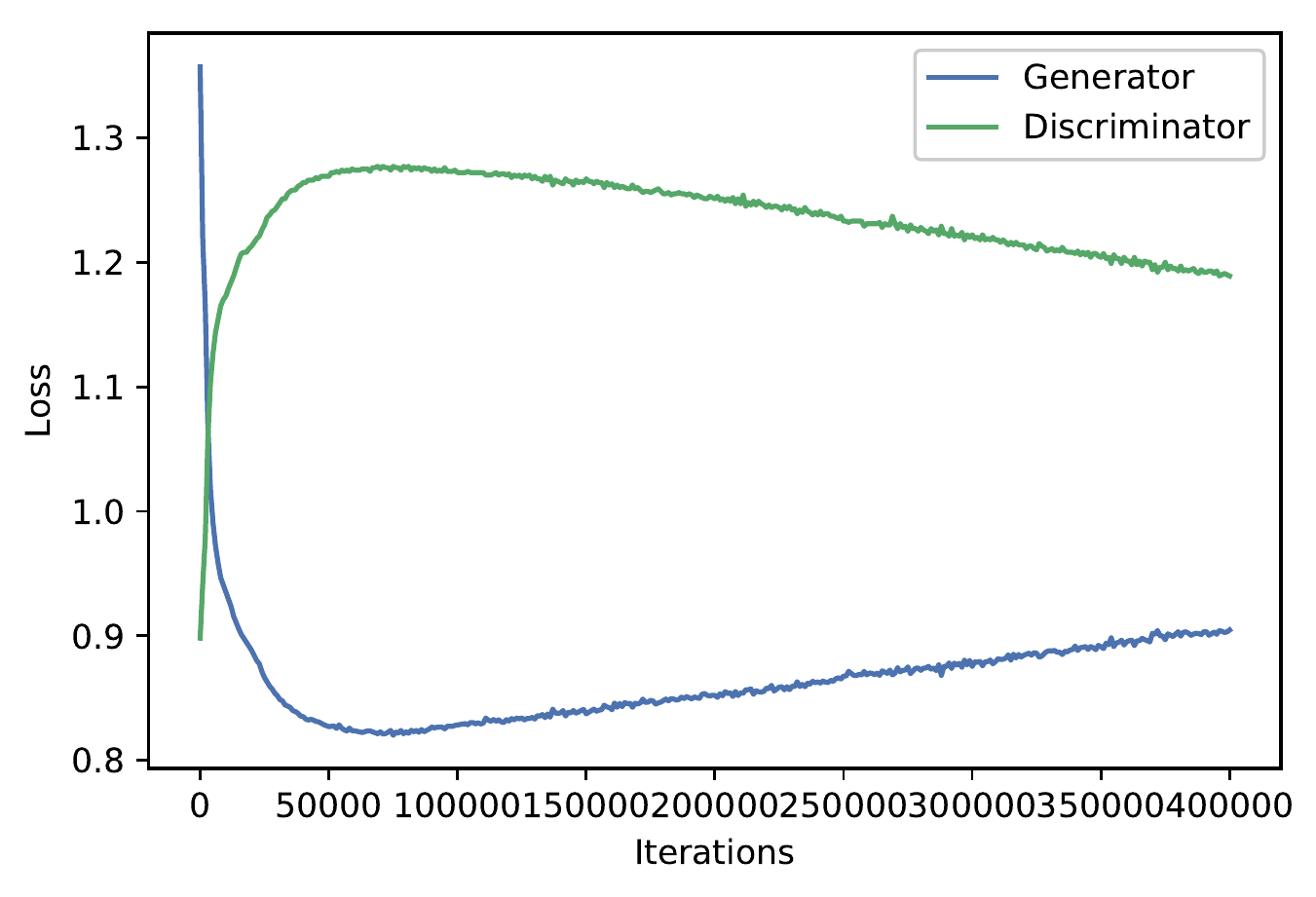}
\caption{Conditional DRAGAN} \label{fig:3b}
\end{subfigure}
\caption{Comparing convergence of conditional GAN and conditional DRAGAN} \label{fig:3}
\end{figure}

\section{Preprocessing Data for GAN Training}
\label{Sec:processing}

Preprocessing of data used in GAN training is one of the crucial steps that determines success or failure of building a GAN that can accurately reproduce original data. In the domain of computer vision, where GANs have been used the most since their introduction, all data features (image pixel values) are numerical, continuous, with similar range and distribution. In many other domains, however, data does not have these nice properties. In particular, financial data is usually comprised of features that have 
\begin{itemize}
  \item different types (numerical continuous, numerical discrete, categorical) 
  \item different ranges
  \item different distributions (including very skewed ones, where the most frequent value occurs in more than 90\% of samples)
  \item missing values (including variables with up to 90\% of missing values)
  \item special values used to denote missing values, cap feature range on the left or on the right, etc.
\end{itemize}
These data properties pose unique challenges for GAN training, which should be addressed to ensure good training and testing performance of GAN.

For our datasets, introduced in Section \ref{Sec:GAN}, we found that the following preprocessing steps allowed good training and subsequently good testing performance of GANs. 
\begin{enumerate}
	\item One-hot encoding for categorical features.
	\item Missing value indicator feature: a new feature that is equal to one in samples where the original feature has missing value, and zero – in all other samples.
	\item Box-Cox transformation \cite{box1964}.
	\item Standard scaling or min-max scaling.
	\item Imputing missing values. 
	
\end{enumerate}

\section{Evaluating data generated by GANs}
\label{Sec:evaluation}
When GANs are used to generate images, the quality of produced samples can be easily evaluated by visual observation: humans can easily determine if generated images look similar to the images from the training set, and are overall of good quality. For non-image data, there are no commonly accepted techniques for evaluating the quality of generated data.

We can start with checking whether histograms of generated features match those of real features. However, matching distributions of individual features and target variable do not guarantee that all existing interactions (relationships) \emph{between features}, as well as \emph{between features and target variable} were also replicated by the generator. Therefore, it is necessary to evaluate and compare overall distribution of generated and real data. Since relationship between features and target variable is the most important one for developing ML models, we can perform a separate test of this relationship by comparing supervised ML models trained on generated and real data. ML models with similar performance would be indicative of good replication of dependencies between features and target variable.

To summarize, we propose that a good generator should produce data that satisfy the following three criteria:
\begin{enumerate}
\item Distributions of individual features in generated data match those in real data
\item Overall distributions of generated and real data match each other
\item Relationship between features and the target variable in real data is replicated in generated data
\end{enumerate}

In the following subsections, we will show how we performed above tests for the data generated by our GANs. 

\subsection{DataQC tool}
\label{feat_hist}

DataQC \cite{love2018} is an internal automated tool developed at American Express for data quality assessment. It allows users to evaluate similarities and differences between two provided datasets. The tool performs a comprehensive set of data quality tests to quickly highlight how one dataset is different from another. These tests include comparison of feature means, rates of missing values, uni- and multivariate distributions, and extreme values. The tool produces detailed findings and quantitive scores for all the tests that it runs.

We used DataQC to evaluate similarity between GAN-generated and real data. In particular, we used it to compare distributions of individual features in generated and real data. As an example, Fig. \ref{fig:4} shows histograms of four selected features in real and generated data for a GAN trained on Dataset C. For all other features we observed similar, very close match between real and generated distributions. This was the case for all  three Datasets. We noted that our GANs were able to reproduce discrete distributions just as good as continues ones, and that for some continuous variables GANs tended to produce slightly smoothed versions of their distributions. We found that smoothing of distributions can be reduced by increasing the number of hidden layers in the generator, or by increasing the number of training iterations.

\begin{figure}[h]
\begin{subfigure}{0.22\textwidth}
\includegraphics[width=\linewidth]{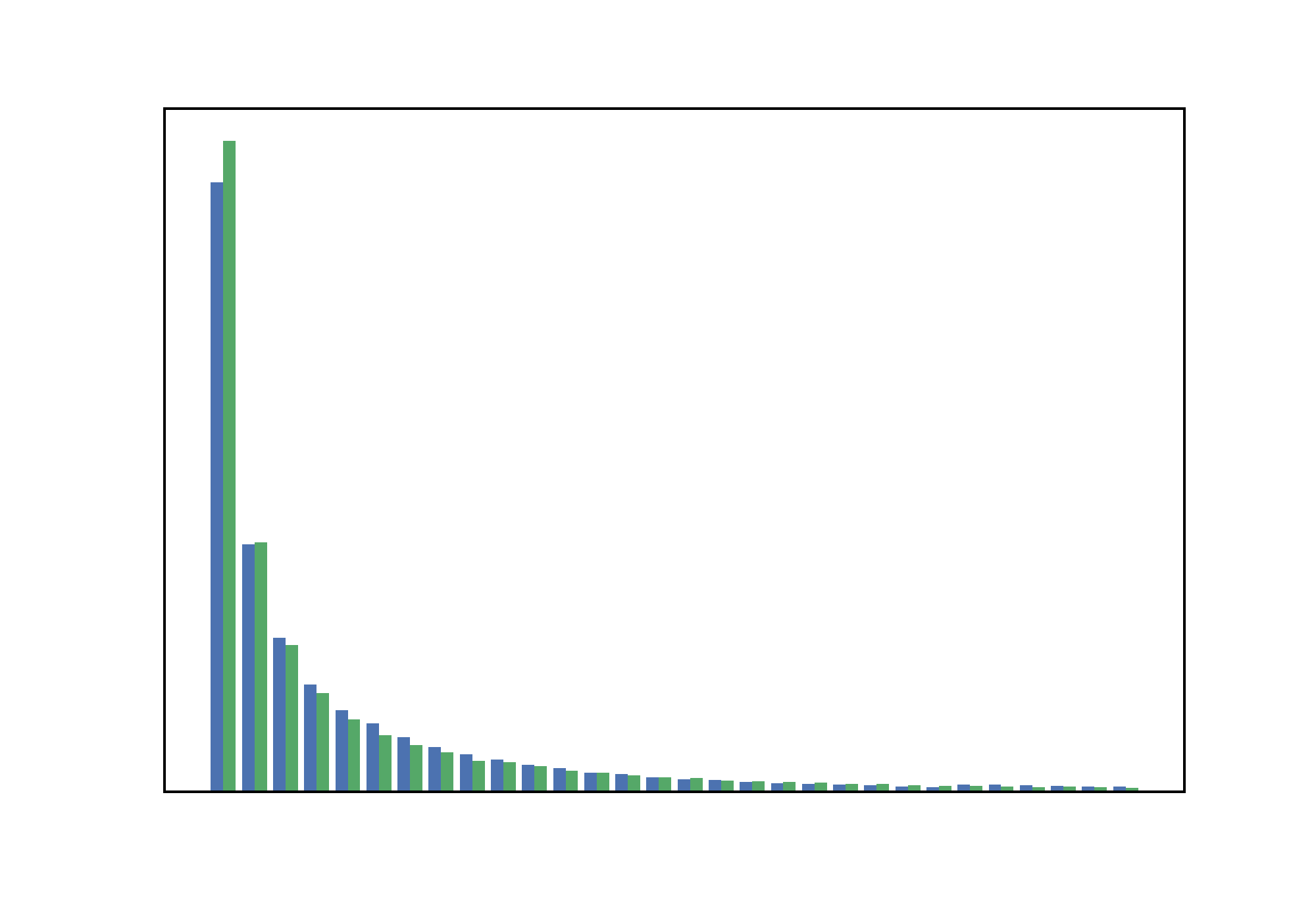}
\caption{Skewed feature} \label{fig:4a}
\end{subfigure}
\hspace*{\fill}
\begin{subfigure}{0.22\textwidth}
\includegraphics[width=\linewidth]{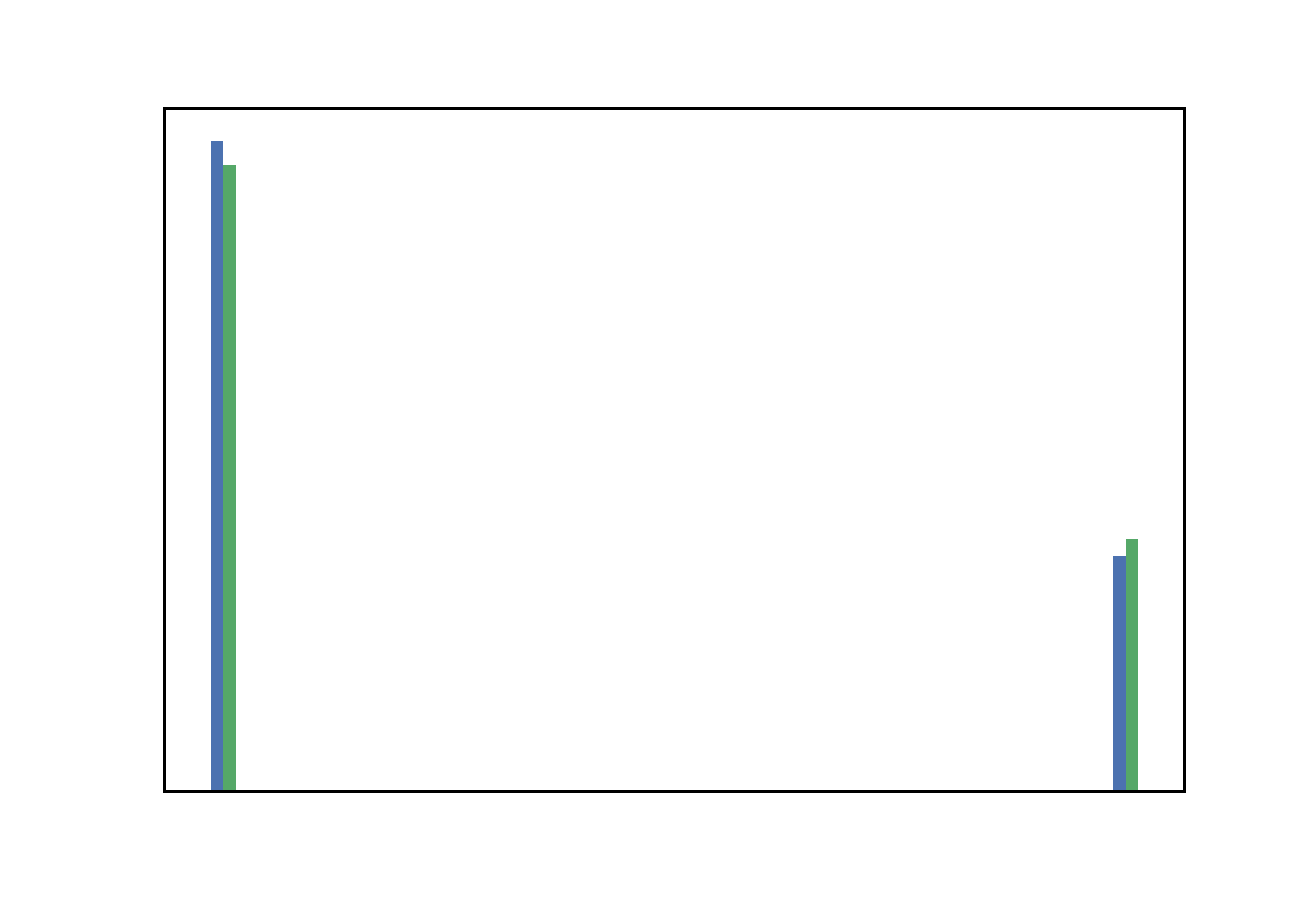}
\caption{Binary feature} \label{fig:4b}
\end{subfigure}
\hspace*{\fill}
\begin{subfigure}{0.22\textwidth}
\includegraphics[width=\linewidth]{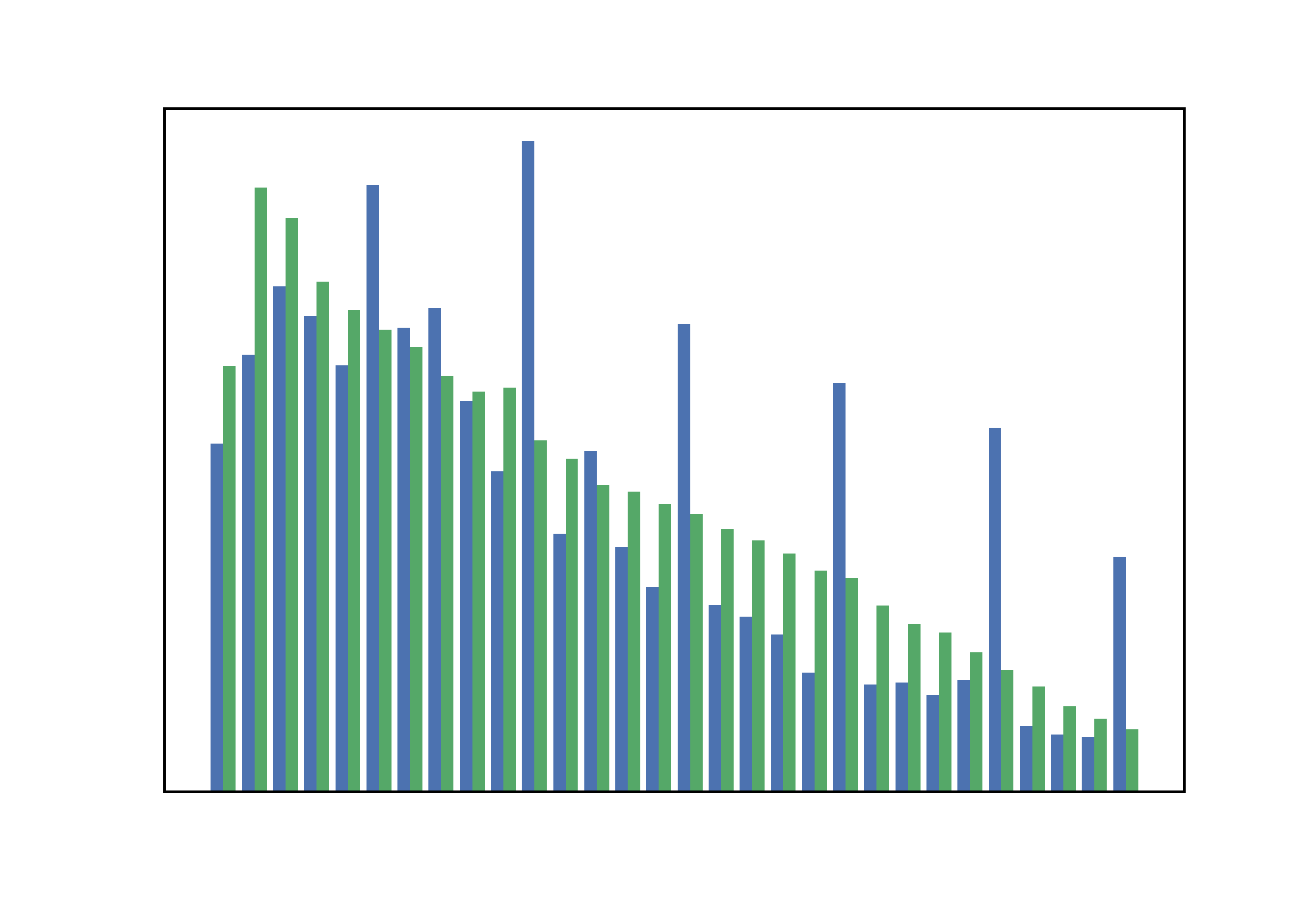}
\caption{Feature with peaks} \label{fig:4c}
\end{subfigure}
\hspace*{\fill}
\begin{subfigure}{0.22\textwidth}
\includegraphics[width=\linewidth]{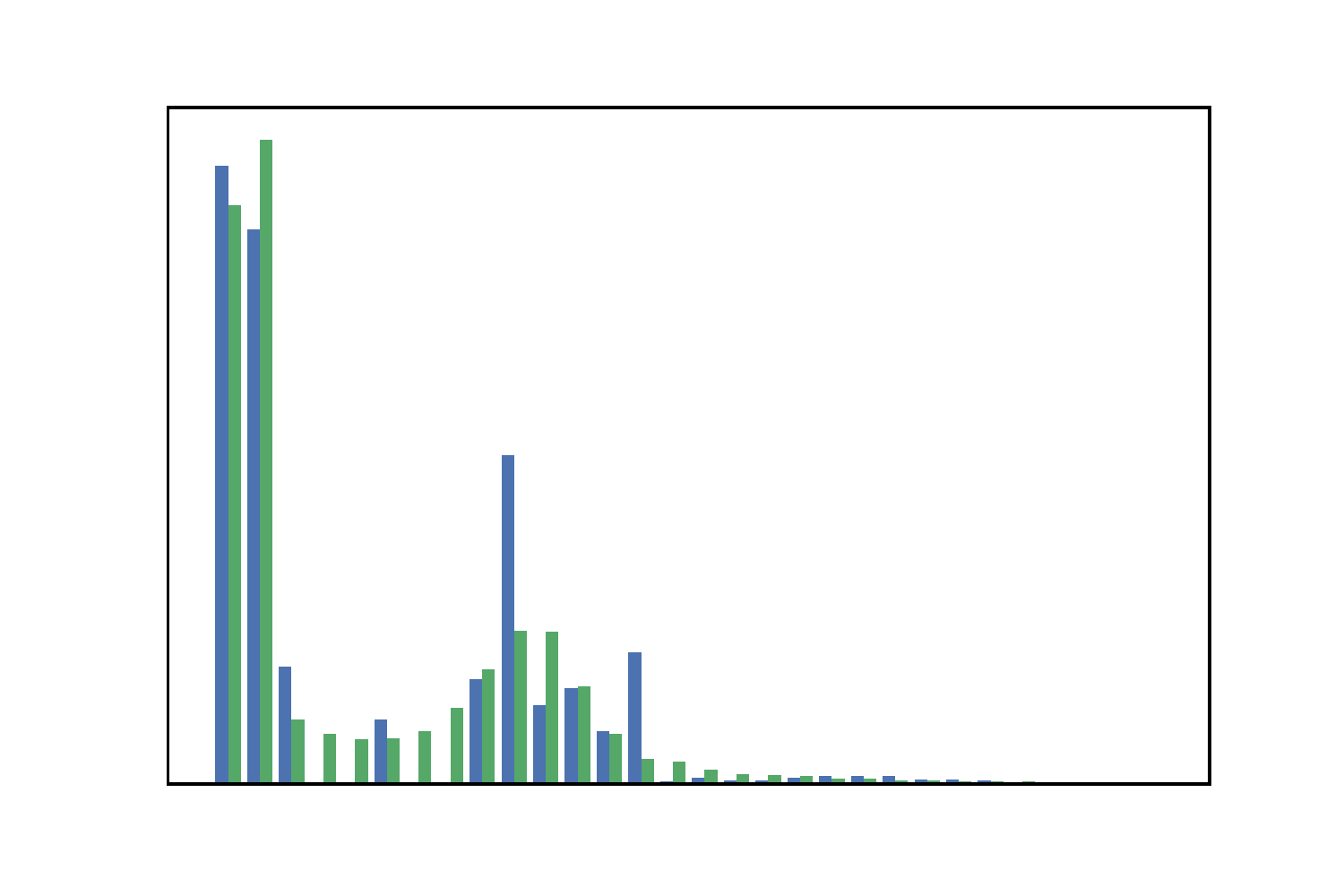}
\caption{Discrete feature} \label{fig:4d}
\end{subfigure}
\caption{Examples of feature histograms for real (blue) and generated (green) data. Experiment with Dataset C.} \label{fig:4}
\end{figure}

\subsection{Visualization with t-SNE algorithm}
\label{tsne_viz}

To compare overall distribution of real and generated data, we visualized the data using t-SNE algorithm \cite{maaten2008}. t-SNE is a transductive algorithm: the model produced by this algorithm cannot be applied to out-of-sample data that was not used to build the model. Therefore, we combined the real and generated data together, obtained t-SNE representation for the combined data, and then split it into two parts corresponding to real and generated data. Fig. \ref{fig:5} shows t-SNE graphs for our experiment with Dataset C. It can be seen in the figure that t-SNE graph for generated data closely matches t-SNE graph for real data, reproducing most of its clusters and gaps with pretty good accuracy in shape and density.

\begin{figure}[h]
\begin{subfigure}{0.5\textwidth}
\includegraphics[width=\linewidth]{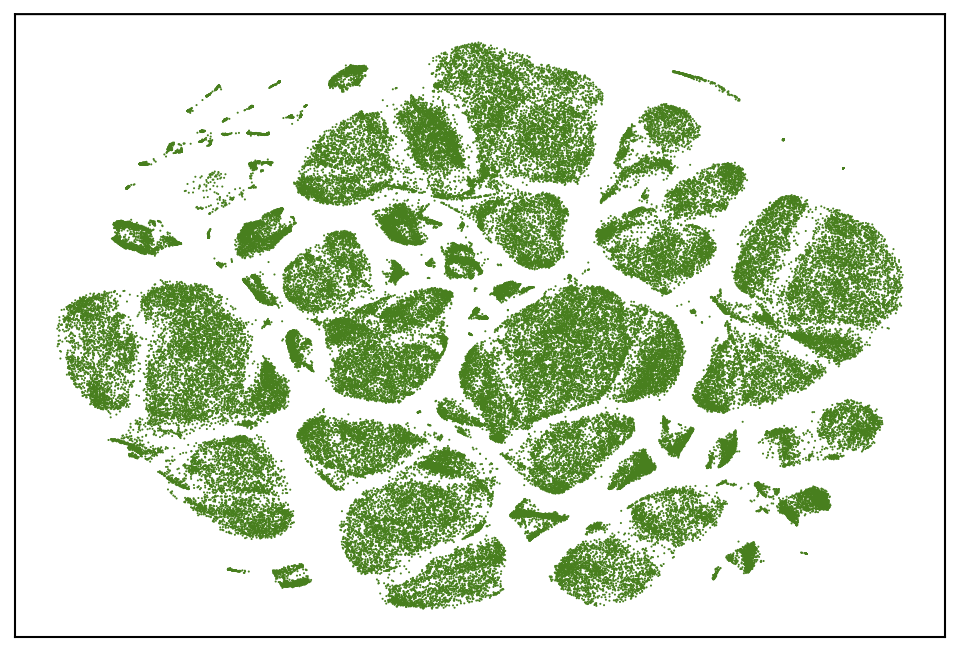}
\caption{Real data} \label{fig:5a}
\end{subfigure}
\hspace*{\fill}
\begin{subfigure}{0.5\textwidth}
\includegraphics[width=\linewidth]{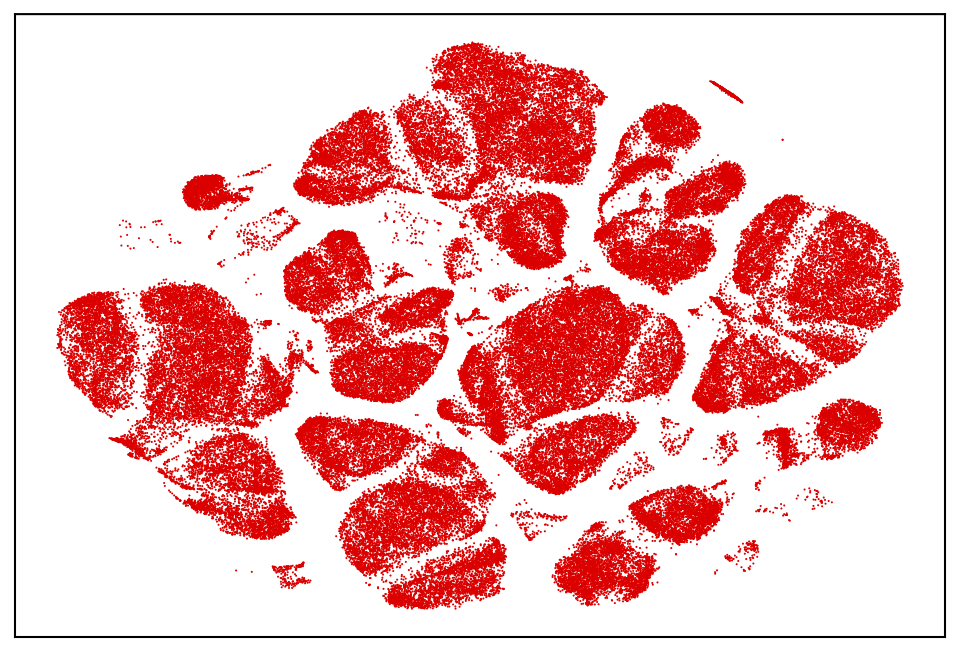}
\caption{Generated data} \label{fig:5b}
\end{subfigure}
\caption{tSNE visualization of real and generated data. Experiment with Dataset C.} \label{fig:5}
\end{figure}

\subsection{Supervised models trained on real and generated data}
\label{super_test}

To test relationship between features and target variable in real and generated data, we compared two supervised ML models: one trained on real data, and another one trained on data produced by GAN (containing the same number of samples as the real data). To train the first model, we used the actual target variable from the real data. To train the second model, we used target variable generated by GAN along with other features. Both ML models used identical hyperparameters during the training phase. Diagram 1 illustrates the idea of this approach.
\begin{table*}[]
  \begin{tabular}{lcccc}
    \hline
    Dataset & Original data & Synthetic data \\
    \hline
 Dataset A & 0.66 & 0.63 \\
 Dataset B & 0.80 & 0.78 \\
 Dataset C & 0.89  & 0.86 \\
 \hline
  \end{tabular}
  \caption{\label{tab:scores} The AUC scores of supervised model test for benchmark datasets.}
\end{table*}

Trained supervised models were validated on an out-of-sample real data with actual values of target variable. Area under the ROC curve (AUC, \cite{powers2011}) was used to compare performance of the two models. AUC scores were calculated for ground truth target values and predictions obtained from the models trained on real and generated datasets; the final AUC scores are presented in Table. \ref{tab:scores}. We found these scores to be close enough to assume that our GAN model replicated the relationship between the target variables and the features with good accuracy.

\begin{figure}[!h]
\centering
\begin{tikzpicture}

 \node[state,
   anchor=center,
  text width=2.3cm] (TD) 
 {Training data};

 \node[state,
   right of=TD,
   anchor=center,
  text width=2.3cm,
  node distance=3.0cm] (GA) 
 {GAN};

 \node[state,
   right of=GA,
   anchor=center,
  text width=2.3cm,
  node distance=3.0cm] (SD) 
 {Synthetic data};

 \node[state,
   below of=GA,
   anchor=center,
  text width=2.3cm,
  node distance=1.2cm] (SA) 
 {Supervised algorithm};

 \node[state,
   below of=TD,
   anchor=center,
  text width=2.3cm,
  node distance=1.2cm] (M1) 
 {Trained model 1};

 \node[state,
   below of=SD,
   anchor=center,
  text width=2.3cm,
  node distance=1.2cm] (M2) 
 {Trained model 2};

 \node[state,
   below of=SA,
   anchor=center,
  text width=2.3cm,
  node distance=1.2cm] (OD) 
 {Out-of-sample validation data};

 \node[state,
   below of=M1,
   anchor=center,
  text width=2.3cm,
  node distance=1.2cm] (TP) 
 {Performance on real data};

 \node[state,
   below of=M2,
   anchor=center,
  text width=2.3cm,
  node distance=1.2cm] (SP) 
 {Performance on synthetic data};

\draw[->] (TD) -- (SA);
\draw[->] (SD) -- (SA);
\draw[->] (TD) -- (GA);
\draw[->] (GA) -- (SD);
\draw[->] (SA) -- (M1);
\draw[->] (SA) -- (M2);
\draw[->] (M1) -- (OD);
\draw[->] (M2) -- (OD);
\draw[->] (OD) -- (TP);
\draw[->] (OD) -- (SP);

\end{tikzpicture}
\caption*{Diagram 1. Supervised model test overview.}
\end{figure}
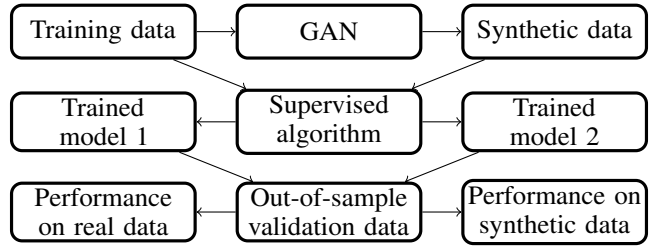

\section{Results and conclusions}
\label{Sec:results}

The final results are presented in Table 2. The models trained on synthetic data show slightly worse performances on out-of-sample validation data but the scores are very close. Generated data can be considered as good approximation of the original data though the performance of the models trained on the original data is still better than on synthetic data. Overall, we conclude that GANs can learn and accurately reproduce intricate features distribution and relationships between features of real modeling data.

We will continue our research in the following  two directions. First, we would like to explore possible causes for lower performance of the models trained on synthetic data. We will also investigate different preprocessing approaches and GAN architectures in order to compare, and see if the baseline performance can be achieved or improved on pure GAN-generated data. We will also work on theoretical justification of the proposed approach, e.g. addressing the following question: given two datasets original and synthetic, what are the sufficient criteria for synthetic data to produce the model with the same performance on original data?

\end{document}